\documentclass[letterpaper, 10 pt, conference]{ieeeconf} 
\IEEEoverridecommandlockouts 
\usepackage{graphicx}
\usepackage{amssymb}
\usepackage{amsmath}
\usepackage{threeparttable}
\usepackage{algorithm}
\usepackage{algorithmic}
\usepackage{cite}
\usepackage{url}
\usepackage{color}

\makeatletter
\let\NAT@parse\undefined

\newcommand{\Rmnum}[1]{\expandafter\@slowromancap\romannumeral #1@}
\makeatother

\usepackage[colorlinks, linkcolor=black, anchorcolor=blue, citecolor=black]{hyperref}

\title
{\LARGE \bf
A New Multi-vehicle Trajectory Generator to Simulate Vehicle-to-Vehicle Encounters
}

\author{Wenhao Ding$^{1}$, Wenshuo Wang$^{2}$, and Ding Zhao$^{2}$
\thanks{$^{1}$Wenhao Ding is with the Department of Electronic Engineering, Tsinghua University, Beijing 10084, China {\tt\small dingwenhao95@gmail.com}}%
\thanks{$^{2}$Wenshuo Wang and Ding Zhao are with the Department of Mechanical Engineering, Carnegie Mellon University (CMU), Pittsburgh, PA 15213, USA. {\tt\small wwsbit@gmail.com}, {\tt\small dingzhao@cmu.edu}}%
}

\begin{document}
\maketitle

\begin{abstract}
Generating multi-vehicle trajectories from existing limited data can provide rich resources for autonomous vehicle development and testing. This paper introduces a multi-vehicle trajectory generator (MTG) that can encode multi-vehicle interaction scenarios (called driving encounters) into an interpretable representation from which new driving encounter scenarios are generated by sampling. The MTG consists of a bi-directional encoder and a multi-branch decoder. A new disentanglement metric is then developed for model analyses and comparisons in terms of model robustness and the independence of the latent codes. Comparison of our proposed MTG with $\beta$-VAE and InfoGAN demonstrates that the MTG has stronger capability to purposely generate rational vehicle-to-vehicle encounters through operating the disentangled latent codes. Thus the MTG could provide more data for engineers and researchers to develop testing and evaluation scenarios for autonomous vehicles.
\end{abstract}

\section{Introduction}

Autonomous driving is being considered as a powerful tool to bring a series of revolutionary changes in human life. However, efficiently interacting with surrounding vehicles in an uncertain environment continue to challenge the deployment of autonomous vehicles because of scenario diversity. Classifying the range of scenarios and separately designing associated appropriate solutions appears to be an alternative to overcome the challenge, but the limited prior knowledge about the complex driving scenarios pose an equally serious problem \cite{56}. Some studies employed deep learning technologies, such as learning controllers via end-to-end neural networks \cite{57} to handle the large amounts of high-quality data without requiring full recovery of the multi-vehicle interactions. Deep learning technologies, however, is limited to scenarios that have never been shown up in the training data set. 

\begin{figure}[t]
\centering
\includegraphics[width=8cm]{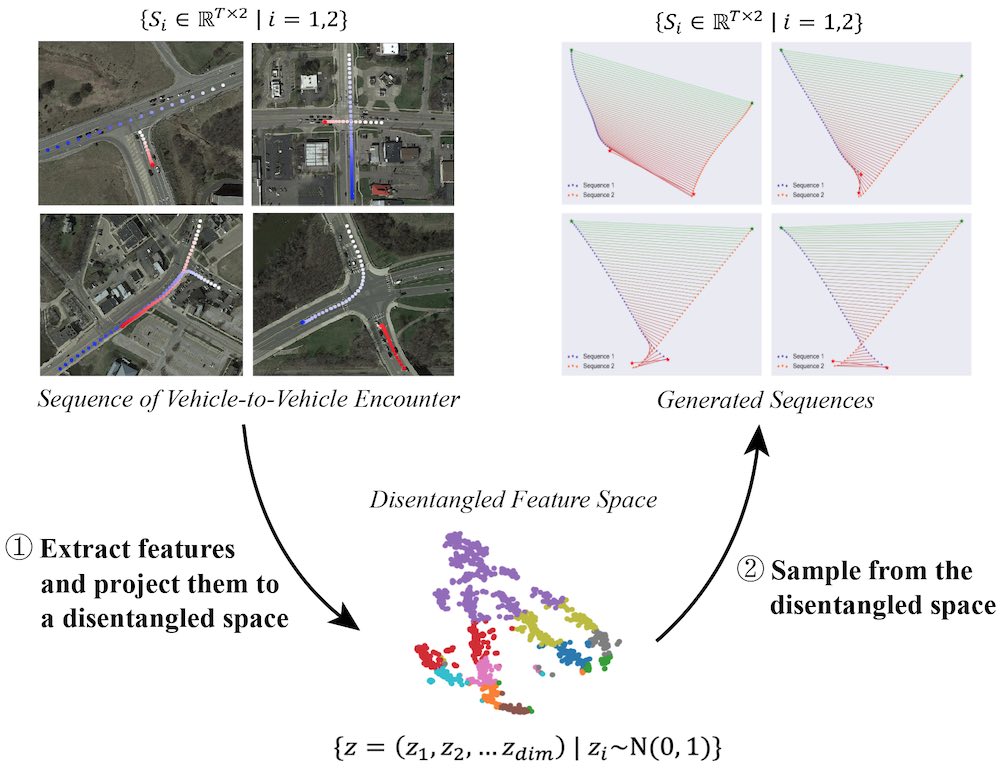}
\caption{Procedure of generating multi-vehicle encounter trajectories.}
\label{fig1}
\end{figure}

Most released databases \cite{33} do not offer sufficient information on multi-vehicle interaction scenarios because of technical limitations and the required costs of collecting data \cite{49}. One alternative is to generate new scenarios that are similar to real world by modeling the limited data available, as shown in Fig.~\ref{fig1}. The basic underlying concept is inspired by image style transformation \cite{14,50} and the functions of deep generative models: projecting the encounter trajectories into a latent space from which new trajectories can be then generated using sampling techniques. Some similar concepts have been developed, for example, variational autoencoders (VAE) \cite{17} which characterize the generated data more explicitly with a Gaussian distribution prior \cite{10}\cite{12}. However, original VAE cannot fully capture the temporal and spatial features of multi-vehicle encounter trajectories because it only handle information over spatial space. As a remedy, the neural networks in a recurrent frame of accounting the history such as long short-term memory (LSTM) \cite{22} and gated recurrent units (GRU) \cite{23} are able to tackle sequences, like vehicle encounter trajectories, ordered by time.

\begin{figure*}[t]
\centering
\includegraphics[width=17cm]{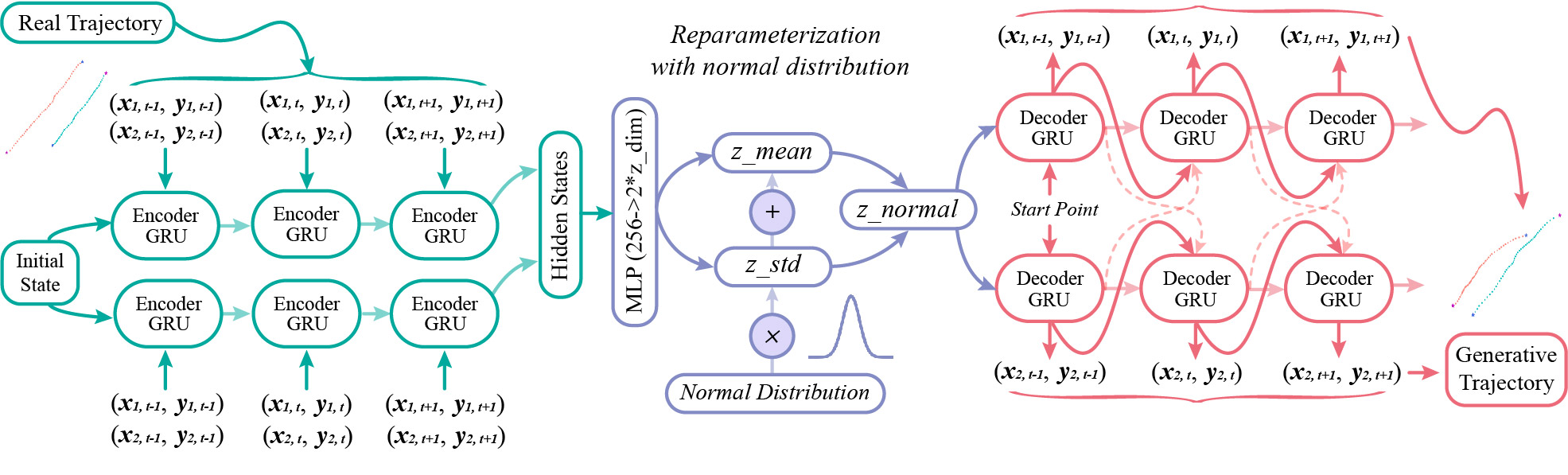}
\caption{Scheme of Multi-Vehicle Trajectory Generator, which consists of three parts: encoder (green), sampling process (purple) and decoder (red).}
\label{fig2}
\end{figure*}

In this paper, we develop a deep generative framework integrated with a GRU module to generate multi-vehicle encounter trajectories. Fig.~\ref{fig2} illustrates our proposed MTG which encodes driving encounters into interpretable representations with a bi-directional GRU module (green), and generates the sequences through a multi-branch decoder (red) separately. The reparameterization process (purple) is introduced between the encoder and the decoder \cite{17}.

On the other hand, model performance evaluation is challenging because of lack of ground truth for the generated multi-vehicle trajectories. Here we propose a new disentanglement metric for model performance evaluation in terms of interpretability and stability. Compared with previous disentanglement metric \cite{12}, our metric is not sensitive to hyper-parameters and can also quantify the relevance among the latent codes.

This paper contributes to the published literature by
\begin{itemize}
\item Introducing a deep generative model that uses latent codes to characterize the dynamic interaction of multi-vehicle trajectories.
\item Generating multi-vehicle trajectories that are consistent with the real data in the spatio-temporal space.
\item Proposing a new disentanglement metric that can comprehensively analyze deep generative models in terms of interpretability and stability.
\end{itemize}

\begin{figure*}[t]
\centering
\includegraphics[width=17cm]{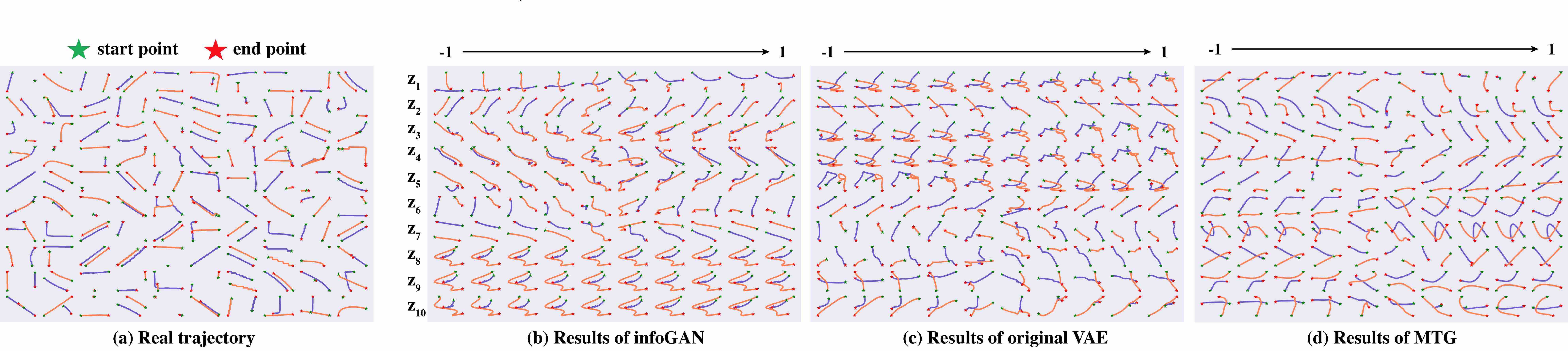}
\caption{Trajectories generated from different models. (a), (b), (c) and (d) represent trajectories of real data, InfoGAN, VAE, and MTG, respectively.}
\label{fig4}
\end{figure*}

\begin{table*}[th]
\begin{center}
\caption{Analysis of Trajectories in the Time Domain}
\label{tab1}
		\begin{tabular}{c||c}
		\hline
		Generated results from MTG with 4 values of $z_1$ (0.1, 0.3, 0.5, 0.7) & Description\\
		\hline
		\begin{minipage}{0.62\textwidth}
   	    \includegraphics[width=105mm, height=15mm]{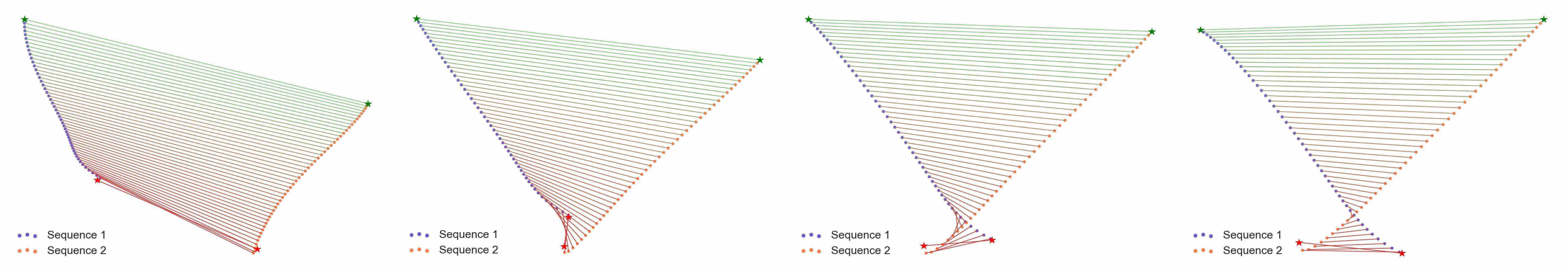}
		\end{minipage} & 
		\begin{minipage}{0.3\textwidth}
		Two vehicles change from driving in the same direction to encountering with each other.
		\end{minipage} \\
		\hline
		\begin{minipage}{0.62\textwidth}
   	    \includegraphics[width=105mm, height=15mm]{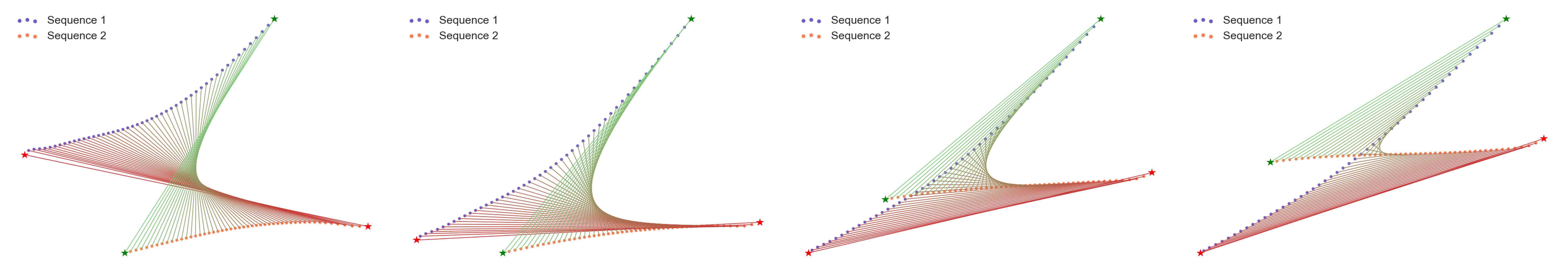}
		\end{minipage} & 
		\begin{minipage}{0.3\textwidth}
		Two vehicles change from driving in the opposite direction to encountering with each other. 
		\end{minipage} \\
		\hline
		\begin{minipage}{0.62\textwidth}
   	    \includegraphics[width=105mm, height=15mm]{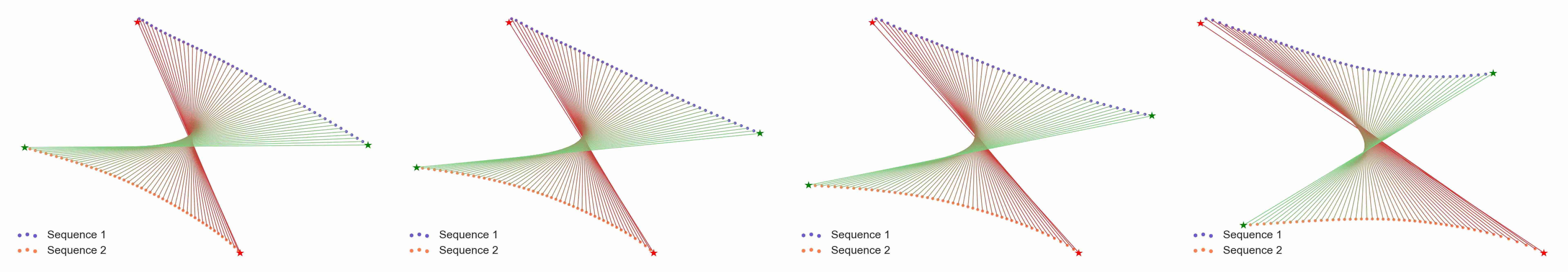}
		\end{minipage} & 
		\begin{minipage}{0.3\textwidth}
		The trajectories of two vehicles rotate in the global coordinate.
		\end{minipage} \\
		\hline
		\end{tabular}
\end{center}
\end{table*}

\section{Related Work}

This section briefly describes four types of deep generative models, an evaluation metric, time series processing methods, and multiple trajectories generation.

\subsection{Deep Generative Models and Evaluation Metric} 

\textbf{Generative Adversarial Networks and InfoGAN.} Generative Adversarial Networks (GAN) consists of two modules\cite{16}: a generator and a discriminator. These two modules compete each other to reach a saddle balance point where the generator can deceive the discriminator by generating high-quality samples. Thus the object function of GAN is formulated as 
\begin{equation}
\begin{split}
&\mathop{\min}_{G}\mathop{\max}_{D}V(G,D) = \\
&\mathbb{{E}}_{x\sim{p}_{data}(x)}[\log D(x)] + {{\mathbb{E}}_{z\sim{p}_{z}(z)}[\log(1-D(G(z)))]}
\end{split} 
\end{equation}
where $x$ represents the real data, $z$ represents the noise, and $G$ and $D$ are the generator and discriminator, respectively. In practice, the inputs of GAN are usually chosen as random noises and thus the output of GAN are unpredictable. In order to generate predictable samples, maximizing the mutual information between the observation and the latent codes is introduced, forming as an InfoGAN  \cite{31}, formulated as
\begin{equation}
\mathop{\min}_{G}\mathop{\max}_{D}[V(G,D) - \lambda I(z; \bar{x} =G(z))]
\end{equation}

\textbf{VAE and $\beta$-VAE.} The optimization of VAE is usually converted to the evidence lower bound (ELBO), since the estimation of the marginal log-likelihood is computationally intractable because of the curse of dimensionality. The optimization function of VAE and $\beta$-VAE is formulated in (3). When $\beta =1$, the original VAE \cite{17} is obtained, and when $\beta>1$, the $\beta$-VAE \cite{10,11} was obtained.
\begin{equation}
\log p_{\theta}(x) = {\mathbb{E}_{q_{\phi}(z|x)} [\log p_{\theta}(x|z)]} - {\beta D_{KL}(q_{\phi}||p(z))}
\end{equation}
where $q_{\phi}$ represents the encoder parameterized by $ \phi$, $p_{\theta}$ represents the decoder parameterized by $\theta$. $\beta$ is used to reduces the distance between $p_{\theta}(z|x)$ and $q_{\phi}(z|x)$, thereby increasing $\beta$ helps obtain more disentanglement. In (3), The first term can be interpreted as a reconstruction error between real and generated trajectories, and the second term is a tightness condition to ensure $p_{\theta}(z|x)=q_{\phi}(z|x)$.

\textbf{Evaluation Metric.} Disentanglement is a key factor for deep generative model evaluation. Some research used the accuracy of classifiers to represent the disentangled ability \cite{10,12}. For example, Higgins \textit{et al.} \cite{10} acquired the training data by calculating the difference $1/L\sum_{l=1}^{L}|z_k^1-z_k^2|$ between the latent codes $z_k^1$ and $z_k^2$ with a fixed $k$ dimension. Kim \textit{et al.} \cite{12} further considered the relationship between the latent variables and the latent codes. The extreme sensitivity of simple classifiers to hyper-parameters, however, can skew the evaluation result. Moreover, the metrics in \cite{10,12} cannot be directly used to analyze the stability and dependency of the latent codes. In Section \uppercase\expandafter{\romannumeral3}-C, we propose a new disentanglement metric capable of comprehensively evaluating model performance without using any classifier-based approaches.

\subsection{Time Series Processing}

Dealing with time series is challenging because of the dependency among adjacent states. Both LSTM \cite{22} and GRU \cite{23} are commonly used to tackle this issue because the forget gate in them allows to control the amount of influence the previous state can lay on the later state. GRU, as a simplified version of LSTM, is more efficient with almost equal performance \cite{37,38}.

Combining RNN/LSTM/GRU and deep generative models has been investigated. In \cite{37}, $\beta$-VAE framework with RNN modules was designed to generate simple sketch drawings with interpretable latent codes. All lines in the sketch were considered as one sequence to avoid the interaction between multiple sequences. The success of \cite{37} partly depends on the use of bi-directional RNN \cite{53}, which extracts more sequential features with a forward-backward data flow \cite{54}.

\subsection{Multiple Trajectories Generation}

Multiple trajectories generation has been used to predict subsequent trajectories under a supervised condition. For example, a multi-LSTM module was used to deal with different sequences and attain concatenated outputs through a tensor \cite{20}. The social LSTM \cite{3} was proposed to generate multi-pedestrian trajectories. The hidden states of different sequences share information to increase the correlation, allowing the agents to interact with their neighborhoods. Later, an improved Social LSTM (or GAN) was proposed  to directly create trajectories from the generator of GAN \cite{5}.

\section{Proposed Methods}

This section will introduce two baselines for comparison, then describe MTG and the new disentanglement metric.

\subsection{Baselines}

In order to to explore the benefits of the modified structure, Baseline 1 is developed using a single-directional GRU encoder of $\beta$-VAE. The encoder processes multiple sequences simultaneously with one GRU module, and the outputs ($\mu$ and $\sigma$) of the encoder are re-sampled through the reparameterization trick \cite{17}. The process is formulated as:
\begin{equation}
{h}_{enc}=GRU_{enc}([S_1; S_2])  
\end{equation}
\begin{equation}
\mu=W_{\mu}h_{enc} + b_{\mu}\ ,\ \sigma=\exp(\frac{W_{\sigma}h_{enc} + b_{\sigma}}{2}) 
\end{equation}
\begin{equation}
z=\mu+\sigma \times \mathcal{N}(0, I)  
\end{equation}
where $S_1$ and $S_2$ are two input sequences (driving encounter), and $z$ is the latent codes in dimension $K$. The decoder takes $z$ as the initial state and outputs sequence coordinates one after another. The two sequences are generated from the decoder at the same time by (7). We select Tanh as the last activation function to make sure the output of the decoder is in [-1,1].
\begin{equation}
[\bar{S_1} ; \bar{S_2}] = GRU_{dec}(P_{start}, z)  
\end{equation}

In order to test another prevalent deep generative framework, Baseline 2 is built on InfoGAN and has the same GRU modules as our MTG. The generator in InfoGAN shares the hidden states among multiple sequences, and the discriminator encompasses a bi-directional GRU. The specific structure of InfoGAN is detailed in Appendix.

\begin{algorithm}[t]
\caption{\textbf{Our disentanglement metric}}
\begin{algorithmic}[1]
\label{alg2}
\STATE $\Sigma=[0.1, 0.4, 0.7, 1.0, 1.3, 1.6, 1.9, 2.2, 2.5, 2.8]$
\STATE initiate $\Omega$ = []
\FOR{$i$ in range(1, $dim(z)$)}
\FOR{$\sigma$ in $\Sigma$}
\STATE initiate $\Theta$ = []
\FOR{$j$ in range(1, $dim(z)$)}
\STATE when $i\ne j$, $z_{j}$ $\sim$ Normal(0, $\sigma$);\
\ENDFOR
\FOR{$l$ in range(1, $L$)}
\STATE $z_i$ $\sim$ Normal(0, $\sigma$);\
\STATE $Z=concat(z_k, k=1\cdots dim(z))$;\
\STATE $Encoder(Z) \Longrightarrow S$,\ $Decoder(S) \Longrightarrow \hat{Z}$;\
\STATE append $\hat{Z}$ to $\Theta$;\
\ENDFOR
\STATE $\omega_{i, \sigma}$ = $Var(\Theta)$,\ append $\omega_{i, \sigma}$ to $\Omega$;\
\ENDFOR
\ENDFOR
\STATE display $\sigma$ for each $i$ in $\Omega$;\
\end{algorithmic}
\end{algorithm}

\subsection{Multi-Vehicle Trajectory Generator (MTG)}

Compared to the Baseline 1, our MTG has two improvements. First, the bi-directional counterparts replace the single-directional GRU module, which enables the encoder to extract deeper representations from the trajectories, because the traffic trajectories are still practically reasonable after being reversed in the time domain. The pipeline of the encoder of MTG is formulated as:
\begin{equation}
{h}_{enc}^{\rightarrow}=GRU_{enc}^{\rightarrow}([S_1; S_2])\ ,\ {h}_{enc}^{\leftarrow}=GRU_{enc}^{\leftarrow}([S_1; S_2])   
\end{equation}
\begin{equation}
h_{enc}=[{h}_{enc}^{\rightarrow}; {h}_{enc}^{\leftarrow}] 
\end{equation}
Second, we separate the decoder into multiple branches and share the hidden states among them. In this way, the hidden state retains all the information over past positions and provides guidance to generate the other sequence. We note that generating two sequences independently avoids mutual influence. The pipeline of the decoder of MTG is formulated as:
\begin{equation}
[S_1^t, h_1^t] = GRU_{dec, 1}(S_1^{t-1}, h_2^{t-1})
\end{equation}
\begin{equation}
[S_2^t, h_2^t] = GRU_{dec, 2}(S_2^{t-1}, h_1^{t-1})
\end{equation}

Then the objective function is concluded as:
\begin{equation}
{\mathcal{L}} = \mathcal{F}(S_1, \bar{S_1}) + \mathcal{F}(S_2, \bar{S_2}) + \beta \times D_{KL}(q_{\phi}||p(z))  
\end{equation}
with $\mathcal{F}(\cdot,\cdot)$ as the mean square error to calculate the reconstruction error, and $\bar{S_i}$ represents the reconstructive trajectory.

\subsection{A New Disentanglement Metric}

The metric in \cite{12} holds one dimension of the latent code fixed and selects other dimensions randomly, then calculates the variance of the output of the encoder under the assumption that the fixed dimension should have the less variance, which is easily recognized by a linear classifier. As a contrast, our metric (see \textbf{Algorithm~\ref{alg2}}) is more stable. We divide the input data into several groups with different variances (each group has $L$ samples) $z_{k, {\sigma}_{m}}$, in which $k\in K$ is the index of the latent code, and $m \in M$ is the group index of the different variances. Only one code was selected and sampled for each group with the others fixed. We input these artificial latent codes $z$ into the decoder to generate the trajectories and then feed them into the encoder to obtain new latent codes $\hat{z}$ again. Finally, we calculate the variances of $\hat{z}$ for each group, revealing the dependence among latent codes and model stability.

\begin{figure}[]
\centering
\includegraphics[width=8.5cm]{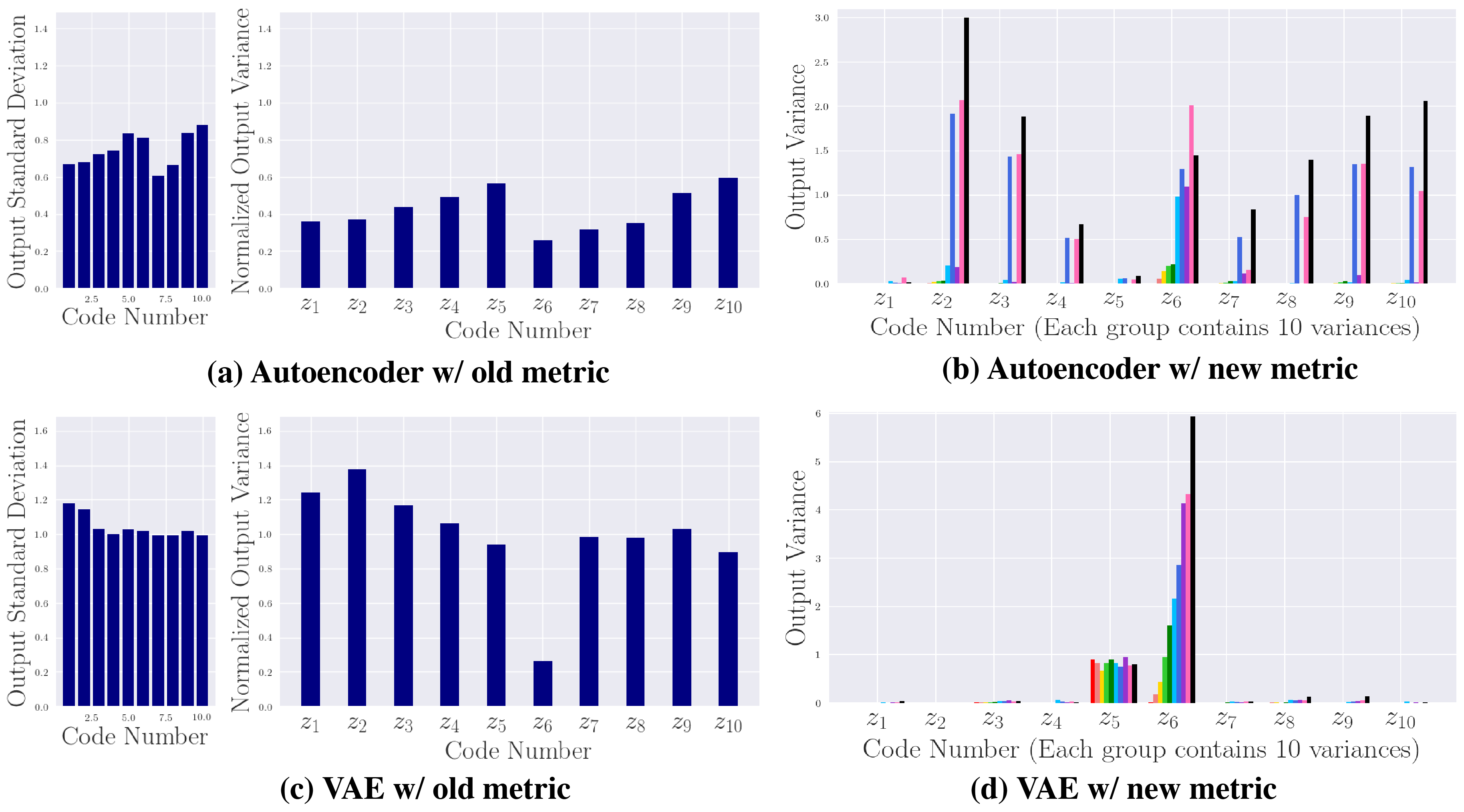}
\caption{Comparison of two evaluation metrics on Autoencoder and VAE.}
\label{fig9}
\end{figure}

\begin{figure*}[t]
\centering
\includegraphics[width=17cm]{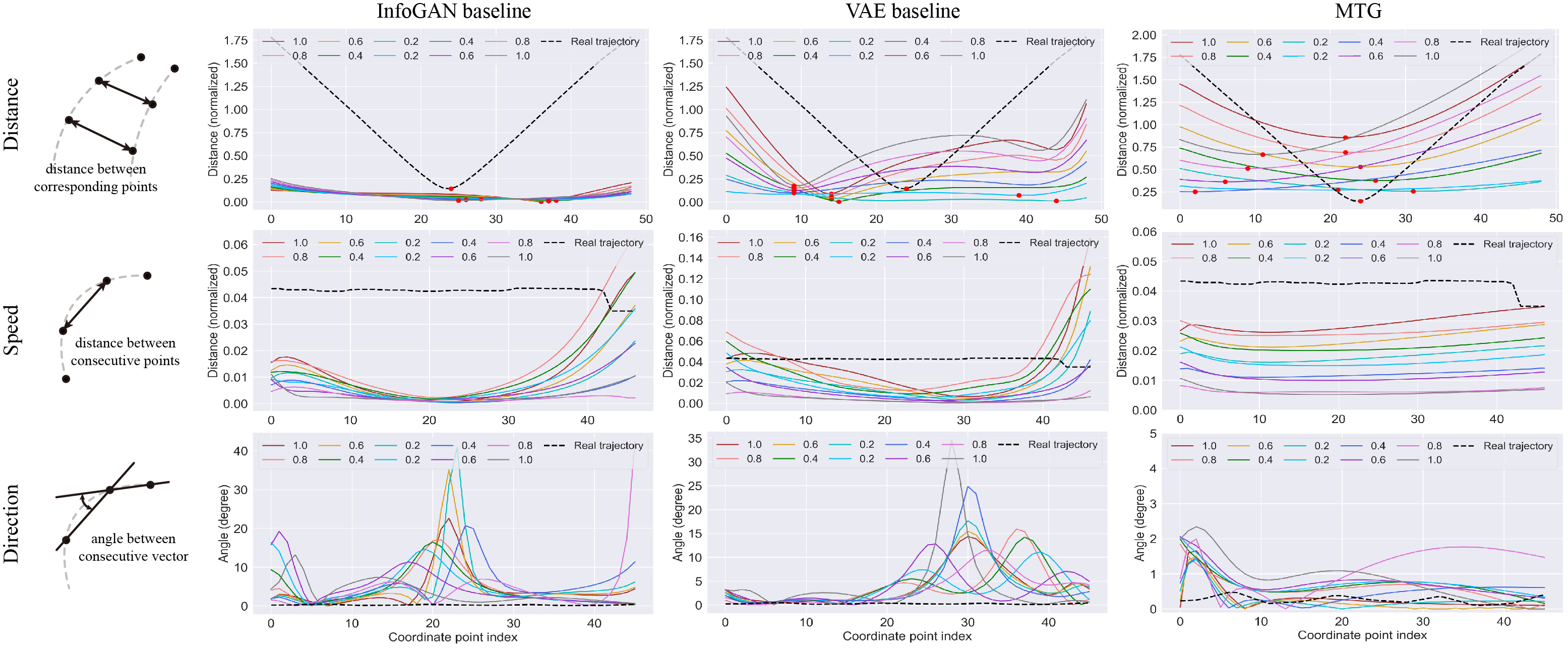}
\caption{Results of three models with traffic rationality evaluation. Both speed and direction results are from one vehicle.}
\label{fig_traffic}
\end{figure*}

\section{Experiments}

\subsection{Dataset and Preprocessing}

We used driving encounter data collected by the University of Michigan Transportation Research Institute (UMTRI) \cite{49} from about 3,500 vehicles equipped with on-board GPS. The latitude and longitude data was recorded by the GPS device to represent vehicle positions. All data was collected with a sampling frequency of 10 Hz. We then linearly upscale and downscale the trajectory vectors to the size of 50. Considering the bias sensitivity of neural networks, we normalize the value of the sequence into the range of [-1, 1] by (13), where $S_i = \{(x_i, y_i) | i=1...50 \}$. 
\begin{equation}
\begin{split}
\{ {\tilde{\mathcal{S}}}_{i} = \frac{{\mathcal{S}}_{i}-\text{mean}({\mathcal{S}}_{i})}{\max({\mathcal{S}}_{1}, {\mathcal{S}}_{2})} | i=1,2\} 
\end{split}
\end{equation}

\subsection{Experiment Settings and Evaluation}

In all experiments, we set the dimension of $z$ to 10. The dimension selection could be different since we do not have prior knowledge about the dimension of the latent space. To test the capability of each code separately, we keep other dimension fixed and change the value of one dimension from -1 to 1 with a step size of 0.1. This step size can be smaller for more detailed analysis. We conduct experiments from two different aspects to compare our MTG with the two baselines.

The first aspect is to evaluate the generative model according to traffic rationality. As shown in the first column of Fig.~\ref{fig_traffic}, we analyze the generated trajectories in time domain with three criteria:
\begin{itemize}
  \item The distance between two sequences, which represents the variation of the distance between two vehicles.
  \item The variation of speed expressed by the distance between two adjacent points.
  \item The variation of trajectory direction for smoothness evaluation, where the angle between two consecutive vectors represents the variation of moving direction.
\end{itemize}

The second aspect is to use our proposed metric to evaluate models in the latent space. For each input variance, we calculate a variance of the output trajectories and display it as a bar as shown in Fig.~\ref{fig5} and Fig.~\ref{fig7}. Each group consists of ten different variances distinguished by color.

\section{Results Analysis}

\subsection{Generative Trajectories Overview}
Fig.~\ref{fig4} shows the generated trajectories from the two baselines and our MTG. Each row shows the variation of one latent code with all others fixed. For the InfoGAN baseline, the last three rows are almost the same, i.e., the codes do not affect any features. This can be explained by the unstable training of InfoGAN. Generating trajectories capable of deceiving the discriminator makes the generator difficult to obtain diversity, since the generator tends to generate similar trajectories that are more likely to mislead the discriminator. As a contrast, the VAE baseline and MTG obtain more diverse trajectories.

Fig.~\ref{fig4} also shows that our MTG attains smoother and more interpretable trajectories (i.e., no circles or sharp turns appear) than the two baselines. The two baselines output some trajectories that are unlikely to appear in the real world. 

Table~\ref{tab1} lists some `zoom-in' figures for more detailed analysis of the generated trajectories of MTG. We connect the associated points in the two sequences, from the starting point (green) to the end point (red), with lines. In each row, the four figures derive from four different values of one latent code with a continuous change from left to right. These trajectories indicate that MTG is able to control some properties of generated trajectories (e.g., the location where two vehicles meet and their directions) through the latent codes.

\subsection{Traffic Rationality Analysis}

Fig.~\ref{fig_traffic} shows some typical generated results based on the three criteria introduced in Section \uppercase\expandafter{\romannumeral4}-B. Different colors represent different values of the latent code $z_1$, wherein black dashed lines represent the real traffic data for comparison.

The corresponding distance indicates that the InfoGAN outputs trajectories with a small variance even for different input values of $z$. This is in line with the problem of mode collapse that is common in GAN family. The VAE baseline and MTG obtain distances closer to the real trajectory. For MTG, the distance gradually decreases and then increases with $z$ changing from -1 to 1, and vehicle speed changes along the latent code value. Comparing the last two vertical columns in Fig.~\ref{fig_traffic} indicates that MTG can generate a much smoother trajectory than VAE. In real world, vehicles cannot take a sharp turning within a very short period of time because of the physical constraints. Therefore, a high value of consecutive angle will reduce the validity.

\subsection{Disentanglement Analysis}

Fig~\ref{fig9}(a) and (b) with $z_6$ and others fixed explain why our metric outperforms previous one \cite{12}. We obtain Fig~\ref{fig9}(a) by using the metric in \cite{12} with an autoencoder. After being normalized by dividing the standard deviation (left plot in Fig~\ref{fig9}(a)), the right part in Fig~\ref{fig9}(a) shows the output variances of $\hat{z}$. Although there is little difference in all codes, $z_6$ still attains the lowest value. Certainly, if all results are close to this case, we can obtain a high accuracy of the classifier while evaluating an autoencoder without any disentangled ability; thus, the metric in \cite{12} has problem evaluating the disentanglement. As a contrast, our metric (Fig~\ref{fig9}(b)) can identify the code that dominates and also reveal the dependency among the codes. High values of latent codes except $z_6$ indicate that a strong dependency among all codes $z_i$.

We then evaluated and compared two metrics on the VAE. Although $z_6$ attains a much low value (Fig~\ref{fig9}(c)) because of the capability of VAE to disentangle the latent code, it is still unsure if the other codes are influenced by $z_6$. Fig~\ref{fig9}(d) shows the nearly zero value of the remaining codes (except $z_5$), i.e. the independence among the codes. Besides proving the disentanglement, we find that $z_5$ is a normal distribution without any information (output variance equal to 1). 

\begin{figure}[t]
\centering
\includegraphics[width=8.5cm]{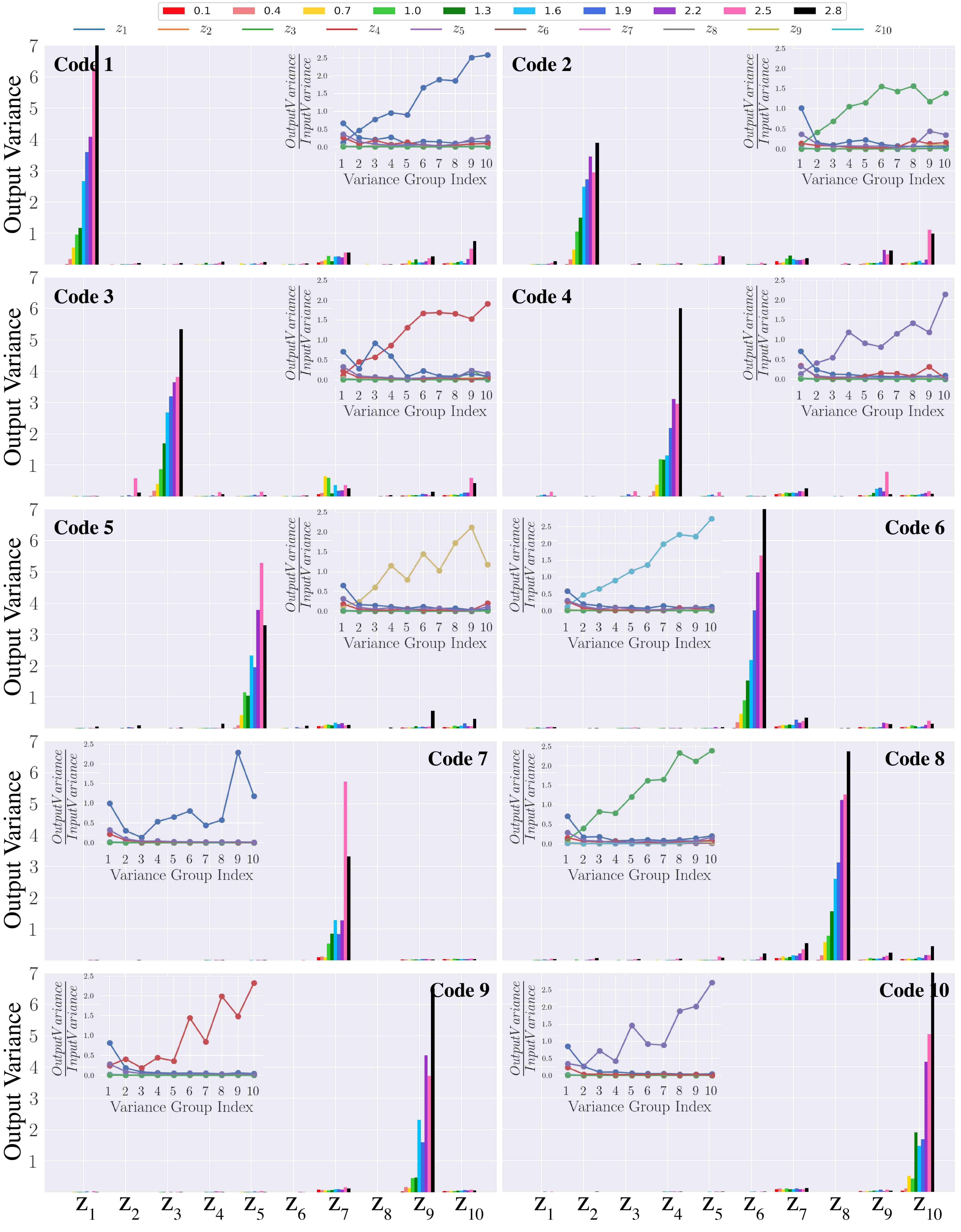}
\caption{Results of MTG with our disentanglement metric.}
\label{fig5}
\end{figure}

\begin{figure}[t]
\centering
\includegraphics[width=8.5cm]{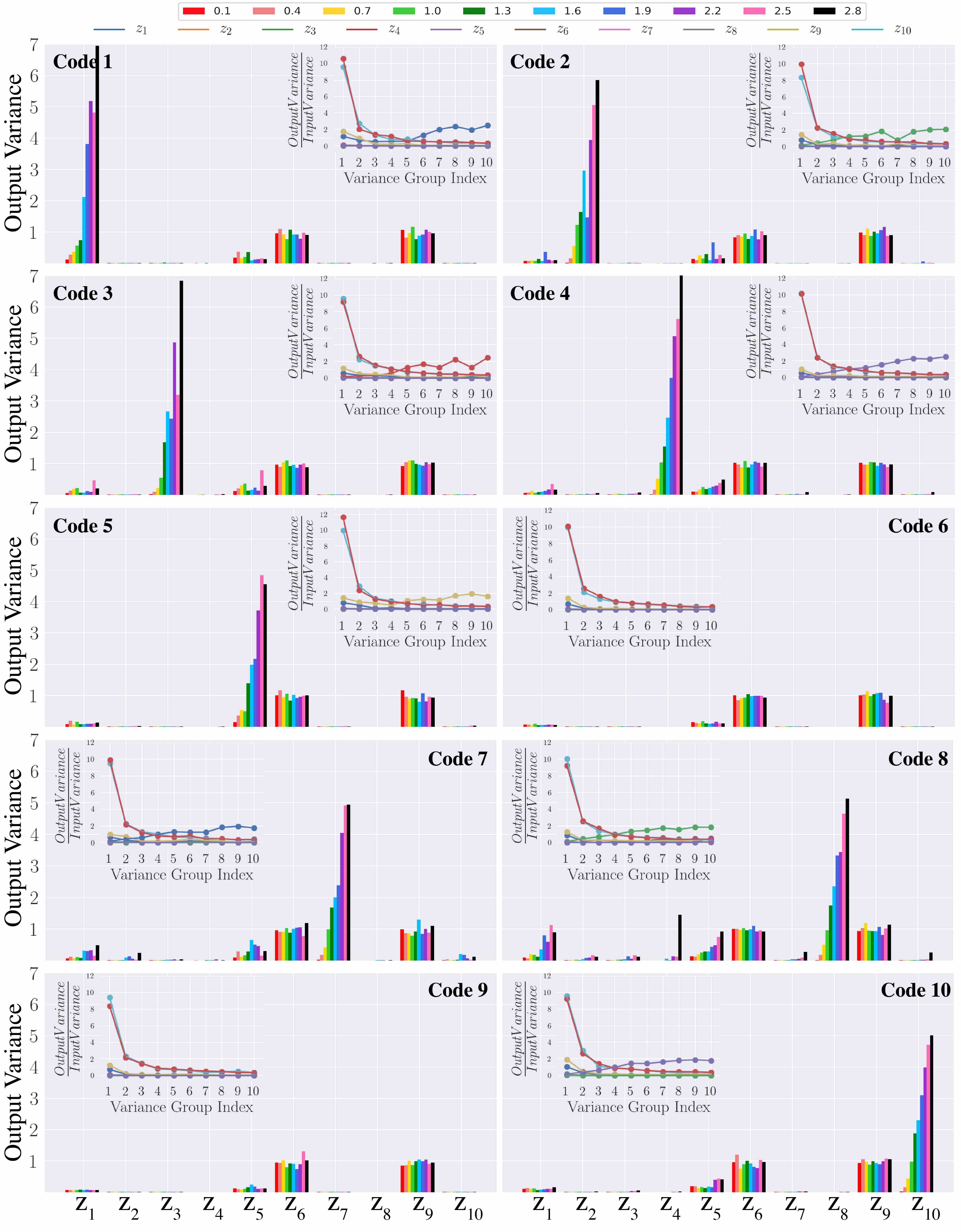}
\caption{Results of VAE baseline with our disentanglement metric.}
\label{fig7}
\end{figure}

At last, we use the proposed metric to compare the VAE baseline with our MTG. Fig.~\ref{fig5} shows (1) that only the sampling code obtains an increasing output variance when increasing the input variance, and (2) that the other codes are close to zero. In other words, there is almost no dependency among the latent codes in MTG, or, changing one code does not influence other features. Fig.~\ref{fig7} shows two normal distributions in the positions of $z_6$ and $z_9$, which indicates that the VAE baseline obtains two latent codes without any useful information. The plot of Code 8 in Fig.~\ref{fig7} also shows that $z_8$ influences $z_1$ and $z_5$ because their output variances are non-zero. 

The subplots inside Fig.~\ref{fig5} and Fig.~\ref{fig7} show the ratio of output variance and input variance. A more robust approach will force the value close to 1. The values in both figures, however, are much greater than 1, which indicates that both the VAE baseline and MTG are not robust enough. A robust generative model requires that the encoder recognize all trajectories generated from the decoder, i.e., the variances of the outputs and the input should be the same.

\section{Conclusion}

Developing safety policies and best practices for autonomous vehicle deployment and low-cost self-driving applications will always depend on the data provided by the high-quality generation of multi-vehicle and pedestrian encounters. This paper proposed a novel method to generate the multi-vehicles trajectories by using publicly available data. Toward the end of extracting the features of two spatiotemporal sequences, a separate generator architecture with shared information was implemented. A new disentanglement metric capable of comprehensively analyzing the generated trajectories and model robustness was also proposed. An evaluation of traffic rationality using the proposed disentanglement metric found that the MTG obtained more stable latent codes and generated high-quality trajectories.

While this paper only considered trajectories with short lengths, the starting point of trajectories can be set arbitrarily and the trajectories cascaded carefully. Future research will account road profiles with conditional deep generative models. The generated encounter data are expected to aid in training automatic driving algorithms and providing individual automatic vehicles with more low-cost data to 'learn' complex traffic scenarios that are rarely encountered in the real world.

\appendix
For hyper-parameter settings, network architectures and more experiment results, please refer to the supplementary material: \url{https://wenhao.pub/publication/trajectory-supplement.pdf}.

\bibliographystyle{IEEEtran}
\bibliography{mybib}

\begin{thebibliography}{10}
\providecommand{\url}[1]{#1}
\csname url@samestyle\endcsname
\providecommand{\newblock}{\relax}
\providecommand{\bibinfo}[2]{#2}
\providecommand{\BIBentrySTDinterwordspacing}{\spaceskip=0pt\relax}
\providecommand{\BIBentryALTinterwordstretchfactor}{4}
\providecommand{\BIBentryALTinterwordspacing}{\spaceskip=\fontdimen2\font plus
\BIBentryALTinterwordstretchfactor\fontdimen3\font minus
  \fontdimen4\font\relax}
\providecommand{\BIBforeignlanguage}[2]{{%
\expandafter\ifx\csname l@#1\endcsname\relax
\typeout{** WARNING: IEEEtran.bst: No hyphenation pattern has been}%
\typeout{** loaded for the language `#1'. Using the pattern for}%
\typeout{** the default language instead.}%
\else
\language=\csname l@#1\endcsname
\fi
#2}}
\providecommand{\BIBdecl}{\relax}
\BIBdecl

\bibitem{56}
T.~Appenzeller, ``The scientists' apprentice,'' \emph{American Association for
  the Advancement of Science}, vol. 357, no. 6346, pp. 16--17, 2017.

\bibitem{57}
S.~Yang, W.~Wang, C.~Liu, and W.~Deng, ``Scene understanding in deep learning
  based end-to-end controllers for autonomous vehicles,'' \emph{IEEE
  Transactions on Systems, Man, and Cybernetics: Systems}, 2018.

\bibitem{33}
W.~Wang and D.~Zhao, ``Extracting traffic primitives directly from
  naturalistically logged data for self-driving applications,'' \emph{IEEE
  Robotics and Automation Letters}, vol.~3, no.~2, pp. 1223--1229, April 2018.

\bibitem{49}
W.~Wang, C.~Liu, and D.~Zhao, ``How much data are enough? a statistical
  approach with case study on longitudinal driving behavior,'' \emph{IEEE
  Transactions on Intelligent Vehicles}, vol.~2, no.~2, p. 85–98, 2017.

\bibitem{16}
I.~Goodfellow, J.~Pouget-Abadie, M.~Mirza, B.~Xu, D.~Warde-Farley, S.~Ozair,
  A.~Courville, and Y.~Bengio, ``Generative adversarial nets,'' in
  \emph{Advances in neural information processing systems}, 2014, pp.
  2672--2680.

\bibitem{14}
M.~Mirza and S.~Osindero, ``Conditional generative adversarial nets,''
  \emph{arXiv preprint arXiv:1411.1784}, 2014.

\bibitem{50}
I.~Gulrajani, F.~Ahmed, M.~Arjovsky, V.~Dumoulin, and A.~C. Courville,
  ``Improved training of wasserstein gans,'' in \emph{Advances in Neural
  Information Processing Systems}, 2017, pp. 5767--5777.

\bibitem{17}
D.~P. Kingma and M.~Welling, ``Auto-encoding variational bayes,'' \emph{arXiv
  preprint arXiv:1312.6114}, 2013.

\bibitem{10}
I.~Higgins, L.~Matthey, A.~Pal, C.~Burgess, X.~Glorot, M.~Botvinick,
  S.~Mohamed, and A.~Lerchner, ``beta-vae: Learning basic visual concepts with
  a constrained variational framework,'' 2016.

\bibitem{12}
H.~Kim and A.~Mnih, ``Disentangling by factorising,'' \emph{arXiv preprint
  arXiv:1802.05983}, 2018.

\bibitem{22}
S.~Hochreiter and J.~Schmidhuber, ``Long short-term memory,'' \emph{Neural
  computation}, vol.~9, no.~8, pp. 1735--1780, 1997.

\bibitem{23}
K.~Cho, B.~Van~Merri{\"e}nboer, C.~Gulcehre, D.~Bahdanau, F.~Bougares,
  H.~Schwenk, and Y.~Bengio, ``Learning phrase representations using rnn
  encoder-decoder for statistical machine translation,'' \emph{arXiv preprint
  arXiv:1406.1078}, 2014.

\bibitem{31}
X.~Chen, Y.~Duan, R.~Houthooft, J.~Schulman, I.~Sutskever, and P.~Abbeel,
  ``Infogan: Interpretable representation learning by information maximizing
  generative adversarial nets,'' in \emph{Advances in neural information
  processing systems}, 2016, pp. 2172--2180.

\bibitem{11}
C.~P. Burgess, I.~Higgins, A.~Pal, L.~Matthey, N.~Watters, G.~Desjardins, and
  A.~Lerchner, ``Understanding disentangling in $beta$-vae,'' \emph{arXiv
  preprint arXiv:1804.03599}, 2018.

\bibitem{37}
D.~Ha and D.~Eck, ``A neural representation of sketch drawings,'' \emph{arXiv
  preprint arXiv:1704.03477}, 2017.

\bibitem{38}
X.~Y. Zhang, F.~Yin, Y.~M. Zhang, C.~L. Liu, and Y.~Bengio, ``Drawing and
  recognizing chinese characters with recurrent neural network,'' \emph{IEEE
  Transactions on Pattern Analysis and Machine Intelligence}, vol.~PP, no.~99,
  pp. 1--1, 2018.

\bibitem{53}
M.~Schuster and K.~K. Paliwal, ``Bidirectional recurrent neural networks,''
  \emph{IEEE Transactions on Signal Processing}, vol.~45, no.~11, pp.
  2673--2681, 1997.

\bibitem{54}
J.~Cross and L.~Huang, ``Incremental parsing with minimal features using
  bi-directional lstm,'' \emph{arXiv preprint arXiv:1606.06406}, 2016.

\bibitem{20}
N.~Deo and M.~M. Trivedi, ``Convolutional social pooling for vehicle trajectory
  prediction,'' \emph{arXiv preprint arXiv:1805.06771}, 2018.

\bibitem{3}
A.~Alahi, K.~Goel, V.~Ramanathan, A.~Robicquet, L.~Fei-Fei, and S.~Savarese,
  ``Social lstm: Human trajectory prediction in crowded spaces,'' in
  \emph{Proceedings of the IEEE Conference on Computer Vision and Pattern
  Recognition}, 2016, pp. 961--971.

\bibitem{5}
A.~Gupta, J.~Johnson, L.~Fei-Fei, S.~Savarese, and A.~Alahi, ``Social gan:
  Socially acceptable trajectories with generative adversarial networks,'' in
  \emph{IEEE Conference on Computer Vision and Pattern Recognition (CVPR)}, no.
  CONF, 2018.

\end{thebibliography}

\end{document}